\documentclass[journal]{IEEEtran}
\usepackage{ifpdf}

\usepackage[noadjust]{cite}

\ifCLASSINFOpdf
   \usepackage[pdftex]{graphicx}
\else
\fi
\usepackage[table,xcdraw]{xcolor}
\usepackage{multirow}

\usepackage[cmex10]{amsmath}
\usepackage{algorithmic}

\usepackage{array}

\usepackage{caption}
\usepackage{subcaption}

\usepackage{fixltx2e}

\usepackage{stfloats}
\usepackage{url}

\begin{document}
%
\title{EddyNet: A Deep Neural Network For Pixel-Wise Classification of Oceanic Eddies}
%
%
%

\author{Redouane~Lguensat,~\IEEEmembership{Member,~IEEE,}
		Miao~Sun, 
        Ronan~Fablet,~\IEEEmembership{Senior~Member,~IEEE,}  
        Evan~Mason, 
        Pierre~Tandeo, 
        and~Ge~Chen
\thanks{R. Lguensat and R. Fablet and P. Tandeo are with IMT Atlantique; UBL; Lab-STICC; 29200 Brest, France. E-mail: redouane.lguensat@imt-atlantique.fr.}
\thanks{M. Sun is with the National Marine Data and Information Service; Key Laboratory of Digital Ocean; 300171 Tianjing, China.}
\thanks{G. Chen is with Department of Marine Information Technology; College of Information Science and Engineering; Ocean University of China; 266000 Qingdao, China.}
\thanks{E. Mason is with the Mediterranean Institute for Advanced Studies (IMEDEA/CSIC-UIB), 07190 Esporles - Balearic Islands, Spain.}

}

%
%

\markboth{Submitted}%
{Lguensat \MakeLowercase{\textit{et al.}}: Bare Demo of IEEEtran.cls for Journals}
%



\maketitle

\begin{abstract}
This work presents EddyNet, a deep learning based architecture for automated eddy detection and classification from Sea Surface Height (SSH) maps provided by the Copernicus Marine and Environment Monitoring Service (CMEMS). EddyNet consists of a convolutional encoder-decoder followed by a pixel-wise classification layer. The output is a map with the same size of the input where pixels have the following labels \{'0': Non eddy, '1': anticyclonic eddy, '2': cyclonic eddy\}. Keras Python code, the training datasets and EddyNet weights files are open-source and freely available on \url{https://github.com/redouanelg/EddyNet}.

\end{abstract}


%
\IEEEpeerreviewmaketitle

\section{Introduction}
%
%
%
%
\IEEEPARstart{G}{oing} "deeper" with artificial neural networks (ANNs) by using more than the original three layers (input, hidden, output) started the so-called deep learning era. The developments and discoveries which are still ongoing are producing impressive results and reaching state-of-the-art performances in various fields. The reader is invited to read \cite{goodfellow2016deep} for a general introduction to deep learning. In particular, Convolutional Neural Networks (CNN) sparked-off the deep learning revolution in the image processing community and are now ubiquitous in computer vision applications. This has led numerous researchers from the remote sensing community to investigate the use of this powerful tool for tasks like object recognition, scene classification, etc... More applications of deep learning for remote sensing data can be found in \cite{zhang2016deep,DLforRS} and references therein.

By standing on the shoulders of recent achievements in deep learning for image segmentation we present "EddyNet", a deep neural network for automated eddy detection and classification from Sea Surface Height (SSH) maps provided by the Copernicus Marine and Environment Monitoring Service (hereinafter denoted by AVISO-SSH). EddyNet is inspired by ideas from widely used image segmentation architectures, in particular U-shaped architectures such as U-Net \cite{ronneberger2015u}. We investigate the use of Scaled Exponential Linear Units (SELU) \cite{klambauer2017self,clevert2015fast} instead of the classical ReLU + Batch Normalization (R+BN) and show that we greatly speed up the training process while reaching comparable results. We adopt a loss function based on the Dice coefficient (also known as the F1 measure) and illustrate that we reach better scores for the two most relevant classes (cyclonic and anticyclonic) than with using the categorical cross-entropy loss. We also supplement dropout layers to our architecture that prevents EddyNet from overfitting.

Our work joins the emerging cross-fertilization between the remote sensing and machine learning communities that is leading to significant contributions in addressing the segmentation of remote sensing images \cite{maggiori,audebert2016semantic,volpi2017dense}. To the best of our knowledge, the present work is the first to propose a deep learning based architecture for pixel-wise classification of eddies, dealing with the challenges of this particular type of data.

This letter is organized as follows: Section \uppercase\expandafter{\romannumeral2} presents the eddy detection and classification problem and related work. Section \uppercase\expandafter{\romannumeral3} describes the data preparation process. Section \uppercase\expandafter{\romannumeral4} presents the architecture of EddyNet and details the training process. Section \uppercase\expandafter{\romannumeral5} reports the different experiments considered in this work and discusses the results. Our conclusion and future work directions are finally stated in Section \uppercase\expandafter{\romannumeral6}. 

\section{Problem statement and related work}

Ocean mesoscale eddies can be defined as rotating water masses, they are omnipresent in the ocean and carry critical information about large-scale ocean circulation \cite{holland1978role,chelton2011global}. Eddies transport different relevant physical quantities such as carbon, heat, phytoplankton, salt, etc. This movement helps in regulating the weather and mixing the ocean \cite{mcwilliams2008nature}. Detecting and studying eddies helps also considering their effects in ocean climate models \cite{le2011parameterization}. With the development of altimeter missions and since the availability of two or more altimeters at the same time, merged products of Sea Surface Height (SSH) reached a sufficient resolution to allow the detection of mesoscale eddies \cite{faghmous2015daily,pascual2006improved}. SSH maps allow us distinguish two classes of eddies: i) anticyclonic eddies that are recognized by their positive SLA (Sea Level Anomaly which is SSH anomaly with regard to a given mean) and ii) cyclonic eddies that are characterized by their negative SLA.
%
%
 
In recent years, several studies were conducted with the aim of detecting and classifying eddies in an automated fashion \cite{faghmous2012eddyscan}. Two major families of methods prevail in the literature, namely, physical parameter-based methods and geometrical contour-based methods. The most popular representative of physical parameter-based methods is the Okubo-Weiss parameter method \cite{okubo1970horizontal,weiss1991dynamics}. The Okubo-Weiss parameter method is however criticized for its expert-based and region-specific parameters and also for its sensitivity to noisy SSH maps \cite{chelton2007global}. Other methods were since then developed using other techniques such as wavelet decomposition \cite{turiel2007wavelet}, winding angle \cite{sadarjoen1999geometric}, etc. Geometric-based methods rely on considering the eddies as elliptic shapes and use closed contour techniques, the most popular method remains Chelton et al. method \cite{chelton2011global} (hereinafter called CSS11). Methods that combines ideas from both worlds are called hybrid methods (e.g. \cite{yi2014enhancing,isern2003identification}). Machine learning methods were also used in the past to propose a solution to the problem \cite{castellani2006identification,hai2008automatic}, recently they are again getting an increasing attention \cite{ashkezari2016oceanic,deepeddy}.  
%
%
%
%

\begin{figure}[t]
\centering
\includegraphics[width=5cm]{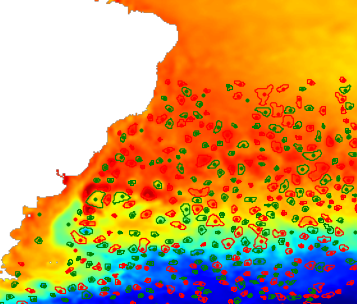}
\caption{A snapshot of a SSH map from the Southern Atlantic Ocean with the detected eddies by PET14 algorithm, red shapes represent anticyclonic eddies while green shapes are cyclonic eddies}
\label{fig:eddies}
\end{figure}

We propose in this work to benefit from the advances in deep learning to address ocean eddy detection and classification. Our proposed deep learning based method requires a training database consisting of SSH maps and their corresponding eddy detection and classification results. In this work, we train our deep learning methods from the results of the \textit{py-eddy-tracker} SSH-based approach (hereinafter PET14) \cite{mason2014new}, the algorithm developed by Mason et al. is closely related to CSS11 but has some significant differences such as not allowing multiple local extremum in an eddy. An example of a PET14 result is given in Figure \ref{fig:eddies} which shows eddies identified in the southwest Atlantic (see \cite{mason17}). 
%
%
The outputs of the eddy tracker algorithm provide the center coordinates of each classified eddy along with its speed and effective contours. Since we aim for a pixelwise classification, i.e., each pixel is classified, we transform the outputs into segmentation maps such as the example shown in Figure \ref{fig:exampletraining}. We consider here the speed contour which corresponds to the closed contour that has the highest mean geostrophic rotational current. The speed contour can be seen as the most energetic part of the eddy and is usually smaller than the effective radius. The next section describes further the data preparation process that yields the training database of pixelwise classification maps.

\section{Data preparation}

As stated in the previous section, we consider PET14 outputs as a training database for our deep-neural-network based algorithms. We use 15 years (1998-2012) of daily detected and classified eddies. The corresponding SSH maps (AVISO-SSH) are provided by the Copernicus Marine Environment Monitoring Service (CMEMS). The resolution of the SSH maps is 0.25$^\circ$.

\begin{figure}[t]
\centering
\includegraphics[scale=0.3]{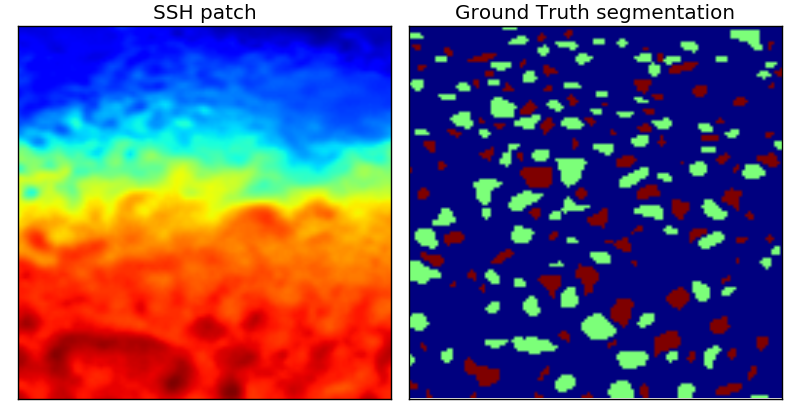}
\caption{Example of a SSH-Segmentation training couple, anticyclonic (green), cyclonic (brown), non eddy (blue)}
\label{fig:exampletraining}
\end{figure}

Due to memory constraints, the input image of our architectures is $128 \times 128$ pixels. The first 14 years are used as a training dataset and the last year (2012) is left aside for testing our architecture. We consider the Southern Atlantic Ocean region depicted in Figure \ref{fig:eddies} and cut the top region where no eddies were detected. Then we randomly sample  one $128 \times 128$ patch from each SSH map, which leaves us with 5100 training samples. A significant property of this type of data is that its dynamics are slow, a single eddy can live for several days or even more than a year. In addition to the fact that a $128 \times 128$ patch can comprise several examples of cyclonic and anticyclonic eddies, we believe that data augmentation (adding rotated versions of the patches to the training database for example) is not needed; we observed experiments (not shown here) that even resulted in performance degradation. The next step consists of extracting the SSH $128 \times 128$ patches from AVISO-SSH. For land pixels or regions with no data we replaced the standard fill value by a zero; this helps to  avoid outliers and does not affect detection since eddies are located in regions with non zero SSH. The final and essential step is the creation of the segmentation masks of the training patches. This is done by creating polygon shapes using the speed contour coordinates mapped onto the nearest lattices in the AVISO-SSH 0.25$^\circ$ grid. Pixels inside each polygon are then labeled with the class of the polygon representing the eddy \{'0': Non eddy/land/no data, '1': anticyclonic eddy, '2': cyclonic eddy\}. An example of the coupled \{SSH map, segmentation map\} from the training dataset is given in Figure \ref{fig:exampletraining}.

\section{Our proposed method}
\subsection{EddyNet architecture}
\begin{figure}[t]
\centering
\includegraphics[scale=0.5]{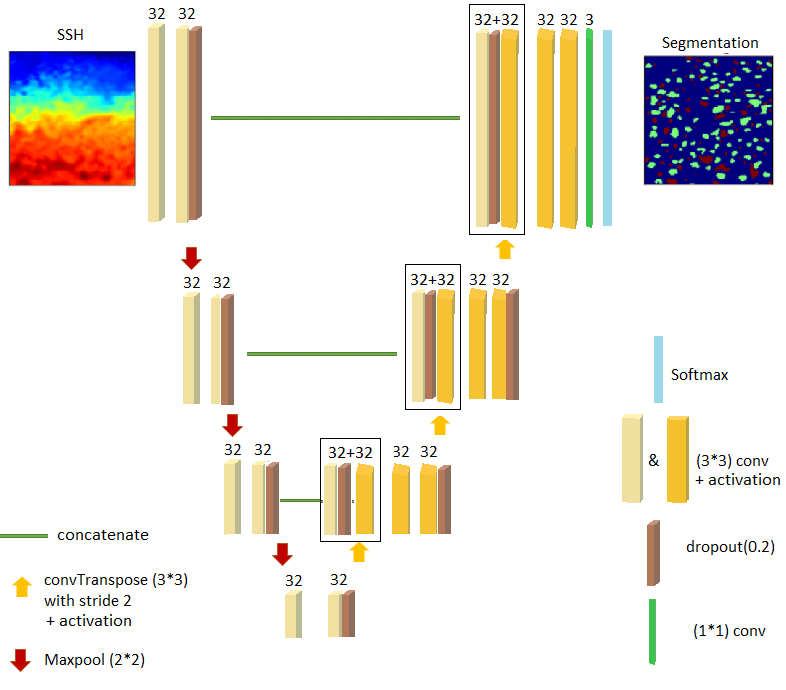}
\caption{EddyNet architecture}
\label{fig:eddynet}
\end{figure}

The EddyNet architecture is based on the U-net architecture \cite{ronneberger2015u}. It starts with an encoding (downsampling) path with 3 stages, where each stage consists of two $3\times 3$ convolutional layers followed by either a Scaled Exponential Linear Unit (SELU) activation function \cite{klambauer2017self} (referred to as EddyNet\_S) or by the classical ReLU activation + Batch Normalization (referred to as EddyNet), then a $2\times 2$ max pooling layer that halves the resolution of the input. The decoding (upsampling) path uses transposed convolutions (also called deconvolutions) \cite{zeiler2010deconvolutional} to return to the original resolution. Like U-net, Eddynet benefits from skip connections from the contracting path to the expanding path to account for information originating from early stages. Preliminary experiments with the original architecture of U-Net showed a severe overfitting given the low number of training samples compared to the capacity of the architecture. Numerous attempts and hyperparameter tuning led us to finally settle on a 3-stage all-32-filter architecture as shown in Figure \ref{fig:eddynet}. EddyNet has the benefit of having a small number of parameters compared to widely used architecture, thus resulting in low memory consumption. Our neural network can still overfit the data which shows that it can capture the nonlinear inverse problem of eddy detection and classification. Hence, we add dropout layers before each max pooling layer and before each transposed convolutional layer; we chose these positions since they are the ones involved in the concatenations where the highest number of filters (64) is present. Dropout layers helped to regularize the network and boosted the validation loss performance. 
%
%
Regarding EddyNet\_S, we mention three essential considerations: i) The weight initialization is different than with EddyNet, we detail this aspect in the experiment section. ii) The theory behind the SELU activation function stands on the self-normalizing property which aims to keep the inputs close to a zero mean and unit variance through the network layers. Classical dropout that randomly sets units to zero could harm this property; \cite{klambauer2017self} propose therefore a new dropout technique called AlphaDropout that addresses this problem by randomly setting activations on the negative saturation value. iii) SELU theory is originally derived for Feed Forward Networks, applying them to CNNs needs careful setting. In preliminary experiments, using our U-net like architecture with SELU activations resulted in a very noisy loss that even explodes sometimes. We think this could be caused by the skip connections that can violate the self-normalizing property desired by the SELU, and hence decided to keep Batch Normalization in EddyNet\_S after each of the maxpooling, transposed convolution and concatenation layers. 

\subsection{Loss metric}
While multiclass classification problems in deep learning are generally trained using the categorical cross-entropy cost function, segmentation problems favor the use of overlap based metrics. The dice coefficient is a popular and largely used cost function in segmentation problems. Considering the predicted region $P$ and the groundtruth region $G$, and by denoting $|P|$ and $|G|$ the sum of elements in each area, the dice coefficient is twice the ratio of the intersection over the sum of areas:
\begin{equation}
\text{DiceCoef}(P,G)=\frac{2 |P \cap G|}{|P|+ |G|}.
\end{equation}
 A perfect segmentation result is given by a dice coefficient of 1, while a dice coefficient of 0 refers to a completely mistaken segmentation. Seeing it from a F1-measure perspective, the dice coefficient is the harmonic mean of the precision and recall metrics.

The implementation uses one-hot encoding vectors, an essential detail is that the loss function of EddyNet uses a soft and differentiable version of the dice coefficient which considers the output of the softmax layer as it is without binarization:
\begin{equation}
\text{softDiceCoef}(P,G)=\frac{2 \sum_i p_i * g_i}{\sum_i p_i+ \sum_i g_i},
\end{equation}
where the $p_i$ are the probabilities given by the softmax layer $0\leq p_i \leq 1$, and the $g_i$ are either 1 for the correct class and 0 either. We found later that a recent study used another version of a soft dice loss \cite{milletari2016v}; a comparison of both versions is out of the scope of this work.

Since we are in the context of a multiclass classification problem, we try to maximize the performance of our network using the mean of three one-vs-all soft dice coefficients of each class. The loss function that our neural network aims to minimize is then simply:
\begin{equation}\label{eq:diceloss}
\text{Dice Loss}= 1 - \text{softMeanDiceCoef}
\end{equation}
\section{Experiments}
\subsection{Assessment of the performance}
Keras framework \cite{chollet2015keras} with a Tensorflow backend is considered in this work. EddyNet is trained on a Nvidia K80 GPU card using ADAM optimizer \cite{kingma2014adam} and mini-batches of 16 maps. The weights were initialized using truncated Gaussian distributed weights of zero mean and \{2/number of input units\} variance \cite{he2015delving} for EddyNet, while we use weights drawn from a truncated Gaussian distribution of zero mean and \{1/number of input units\} variance for EddyNet\_S. The training dataset is split into 4080 images for training and 1020 for validation. We also use an early-stopping strategy to stop the learning process when the validation dataset loss stops improving in five consecutive epochs. EddyNet weights are then the ones resulting in the lowest validation loss value.

\begin{figure}
    \centering
    \begin{subfigure}[b]{0.3\textwidth}
       \centering \includegraphics[width=\textwidth]{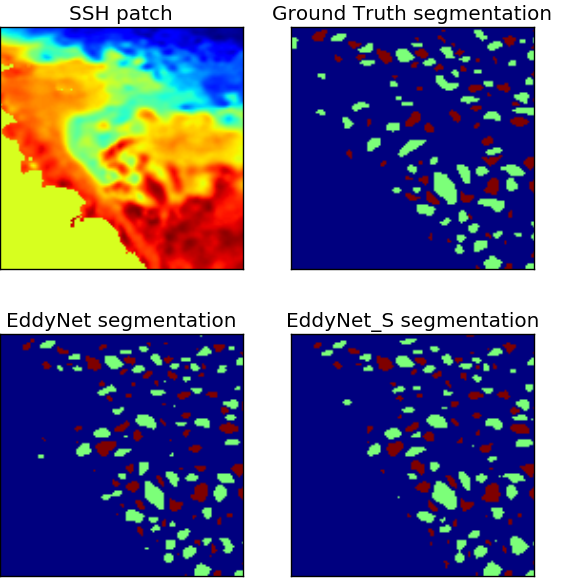}
       \caption{}\label{fig:example1}
    \end{subfigure}
\begin{subfigure}[b]{0.3\textwidth}
       \centering 
       \includegraphics[width=\textwidth]{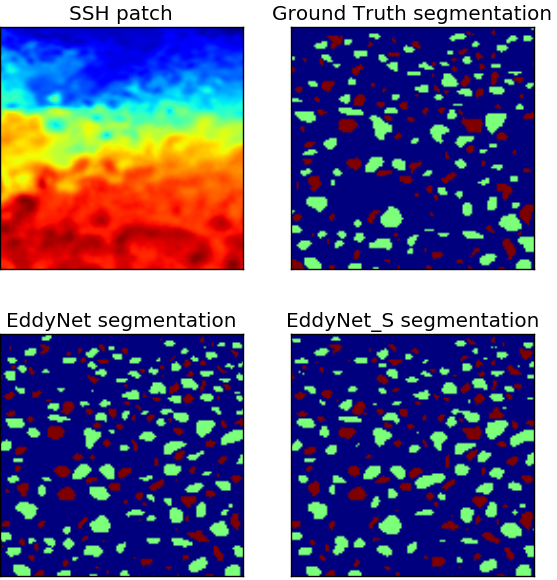}
       \caption{}\label{fig:example2}
    \end{subfigure}
\caption{Examples of the eddy segmentation results using Eddynet and EddyNet\_S: anticyclonic eddies (green), cyclonic (brown), non eddy (blue)}
\label{fig:exampleeddynet}
\end{figure}

EddyNet and EddyNet\_S are then compared regarding the use of the classical ReLU+BN and the use of SELU. We also compare the use of overlap based metric represented by the Dice Loss (Equation \ref{eq:diceloss}), with the classical Categorical Cross-Entropy (CCE). Table \ref{tab:eddynet_results} compares the four combination in terms of global accuracy and mean dice coefficient (original not soft) averaged on 50 random sets of 360 SSH $120\times 120$ maps from 2012. Training EddyNet\_S takes nearly half the time needed for training EddyNet. Comparison regarding the training loss function shows that training with the dice loss results in a higher dice coefficient for our two classes of interest (cyclonic and anticyclonic) in both EddyNet and EddyNet\_S; dice loss yields a better overall mean dice coefficient than training with CCE loss. Regarding the effect of the activation function, we obtained better metrics with EddyNet at the cost of a longer training procedure. Visually Eddynet and EddyNet\_S give close outputs as can be seen in Figure \ref{fig:exampleeddynet}.

\begin{table*}[t]
\centering
\caption{Metrics calculated from the results of 50 random sets of 360 SSH patches from the test dataset, we report the mean value and put the standard variation between parenthesis.}
\label{tab:eddynet_results}
\begin{tabular}{llll|l|l|l|ll}
\cline{5-7}
                                                        &                                         &                                         &                                        & Anticyclonic             & Cyclonic                 & Non Eddy                 &                                               &                                               \\ \cline{2-9} 
\multicolumn{1}{l|}{}                                   & \multicolumn{1}{l|}{\#Param}       & \multicolumn{1}{l|}{Epoch time}     & Train loss                          & \multicolumn{3}{c|}{Dice Coef}                                                & \multicolumn{1}{l|}{Mean Dice Coef}          & \multicolumn{1}{l|}{Global Accuracy}          \\ \hline
\multicolumn{1}{|l|}{}                                  & \multicolumn{1}{l|}{177,571}                   & \multicolumn{1}{l|}{}                   & \cellcolor[HTML]{FFCE93}Dice Loss      & \cellcolor[HTML]{FFCE93}\textbf{0.708} (0.002) & \cellcolor[HTML]{FFCE93}\textbf{0.677} (0.001) & \cellcolor[HTML]{FFCE93}0.929 (0.001) & \multicolumn{1}{l|}{\cellcolor[HTML]{FFCE93}\textbf{0.772 }(0.001)} & \multicolumn{1}{l|}{\cellcolor[HTML]{FFCE93}88.60\% (0.10\%)} \\ \cline{4-9} 
\multicolumn{1}{|l|}{\multirow{-2}{*}{EddyNet}} & \multicolumn{1}{l|}{}                   & \multicolumn{1}{l|}{\multirow{-2}{*}{$\sim$12 min}} & \cellcolor[HTML]{FFFFC7}CCE & \cellcolor[HTML]{FFFFC7}0.695 (0.003) & \cellcolor[HTML]{FFFFC7}0.651 (0.001) & \cellcolor[HTML]{FFFFC7}0.940 (0.001) & \multicolumn{1}{l|}{\cellcolor[HTML]{FFFFC7}0.762 (0.001)} & \multicolumn{1}{l|}{\cellcolor[HTML]{FFFFC7}89.92\% (0.07\%)} \\ \cline{1-1} \cline{3-9} 
\multicolumn{1}{|l|}{}                                  & \multicolumn{1}{l|}{}                   & \multicolumn{1}{l|}{$\sim$7 min}                   & \cellcolor[HTML]{FFCE93}Dice Loss      & \cellcolor[HTML]{FFCE93}0.694 (0.003) & \cellcolor[HTML]{FFCE93}0.665 (0.001) & \cellcolor[HTML]{FFCE93}0.933 (0.001) & \multicolumn{1}{l|}{\cellcolor[HTML]{FFCE93}0.764 (0.001)} & \multicolumn{1}{l|}{\cellcolor[HTML]{FFCE93}88.98\% (0.09\%)} \\ \cline{4-9} 
\multicolumn{1}{|l|}{\multirow{-2}{*}{EddyNet\_S}}    & \multicolumn{1}{l|}{\multirow{-4}{*}{}} & \multicolumn{1}{l|}{\multirow{-2}{*}{}} & \cellcolor[HTML]{FFFFC7}CCE & \cellcolor[HTML]{FFFFC7}0.682 (0.002) & \cellcolor[HTML]{FFFFC7}0.653 (0.002) & \cellcolor[HTML]{FFFFC7}0.939 (0.001) & \multicolumn{1}{l|}{\cellcolor[HTML]{FFFFC7}0.758 (0.001)} & \multicolumn{1}{l|}{\cellcolor[HTML]{FFFFC7}89.83\% (0.08\%)} \\ \hline
\end{tabular}
\end{table*}

\subsection{Ghost eddies}
\begin{figure}
    \centering
    \begin{subfigure}[b]{0.5\textwidth}
       \includegraphics[width=\textwidth]{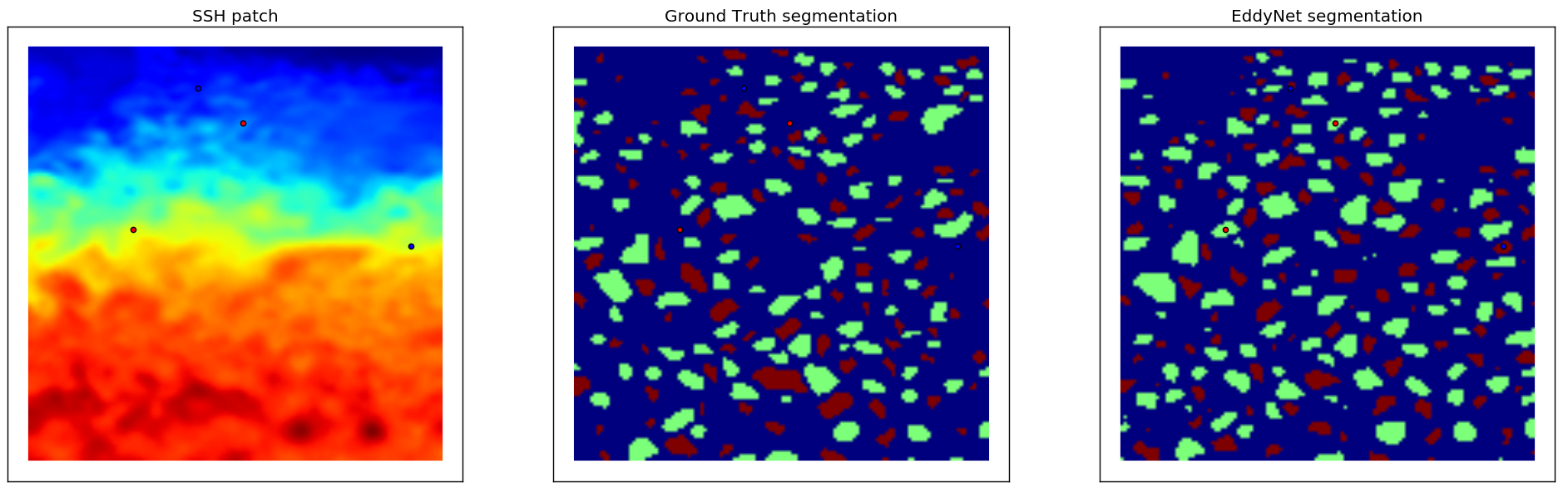}
       \caption{}\label{fig:ghost1}
    \end{subfigure}
    ~ 
    \begin{subfigure}[b]{0.5\textwidth}
       \includegraphics[width=\textwidth]{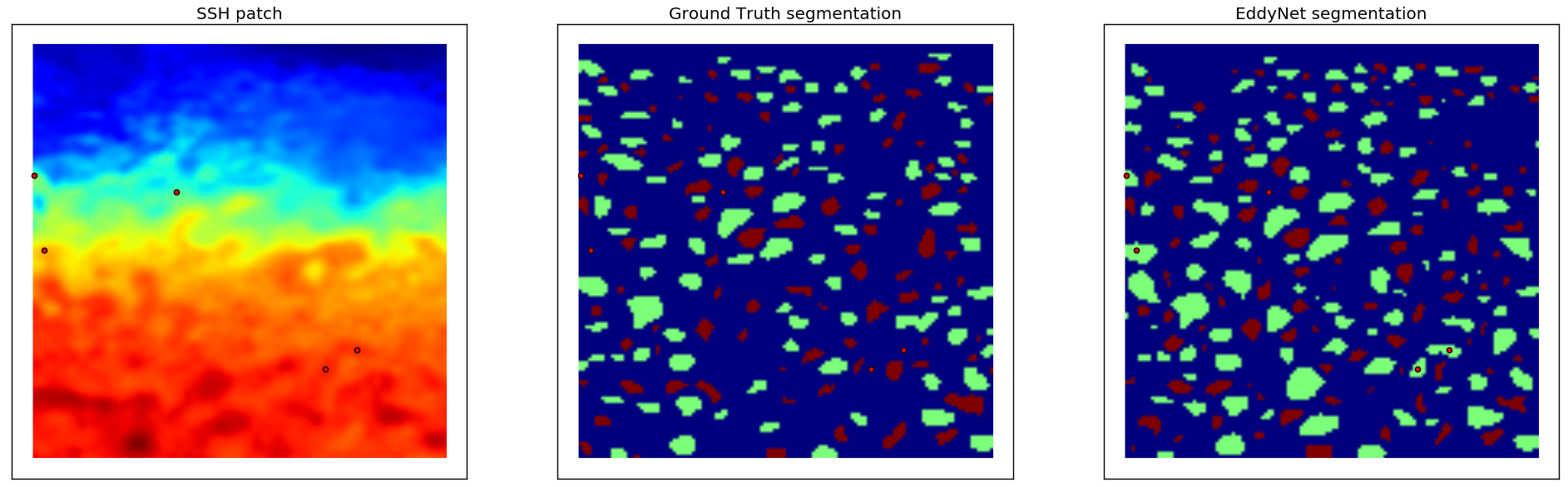}
       \caption{}\label{fig:ghost2}
    \end{subfigure}
\caption{Detection of ghost eddies: [left] SSH map where ghost eddies centers are marked: anticyclonic (red dots), cyclonic (blue dots). [center] PET14 segmentation. [right] EddyNet segmentation: anticyclonic (green), cyclonic (brown), non eddy (blue)}
\label{fig:ghost}
\end{figure}

The presence of ghost eddies is a frequent problem encountered in eddy detection and tracking algorithms \cite{faghmous2015daily}. Ghost eddies are eddies that are found by the detection algorithm then disappear between consecutive maps before reappearing again. To point out the position of the missed ghost eddies, PET14 uses linear temporal interpolation between centers of detected eddies and stores the positions of the centers of ghost eddies. Using EddyNet we check if the pixels of ghost eddy centers correspond to actual eddy detections. We found that EddyNet assigns the centers of ghost eddies to the correct eddy classes 55\% of the time for anticyclonic eddies, and 45\% for cyclonic eddies. EddyNet could be a relevant method to detect ghost eddies that are missed out by conventional methods. Figure \ref{fig:ghost} illustrates two examples of ghost eddy detection. 

\section{Conclusion}
This work investigates the use of recent developments in deep learning based image segmentation for an ocean remote sensing problem, namely, eddy detection and classification from Sea Surface Height (SSH) maps. We propose EddyNet, a deep neural network architecture inspired from architectures and ideas widely adopted in the computer vision community. We transfer successfully the knowledge gained to the problem of eddy classification by dealing with various challenges. Future work involves investigating the use of temporal volumes of SSH and deriving a 3D version inspired by the works of \cite{milletari2016v}. Adding other surface information such as Sea Surface Temperature might also help improving the detection. Another extension would be the application of EddyNet over the globe, and assessing its general capacity over other regions. Post-processing by constraining the eddies to verify additional criteria and tracking the eddies was omitted in this work and could also be developed in future work.  

Beyond the illustrative aspect of this contribution, we offer to the oceanic remote sensing community an easy and powerful tool that can save handcrafting model efforts. Any user can employ his own eddy segmentation "ground truth" and train the model from scratch if he/she has the necessary memory and computing resources, or simply use EddyNet provided weights as an initialization then perform fine-tuning using his/her dataset. One can also think of averaging results from classical contour-based methods and EddyNet. In the spirit of reproducibility, Python code is available at \url{https://github.com/redouanelg/eddynet}, and we also share the training and testing data used for this work to encourage competing methods and, especially, other deep learning architectures.

\section*{Acknowledgment}

The authors would like to thank Antoine Delepoulle, Bertrand Chapron and Julien Le Sommer for their constructive comments. This  work  was  supported  by  ANR (Agence Nationale de la Recherche, grant ANR-13-MONU-0014) and Labex Cominlabs (grant  SEACS). Evan Mason is supported by the Copernicus Marine Environment Monitoring Service (CMEMS) MedSUB project.

\ifCLASSOPTIONcaptionsoff
  \newpage
\fi



\bibliographystyle{IEEEtran}
\bibliography{AnDA_biblio,IEEEabrv,bibtex/bib/IEEEexample}

\begin{thebibliography}{10}
\providecommand{\url}[1]{#1}
\csname url@samestyle\endcsname
\providecommand{\newblock}{\relax}
\providecommand{\bibinfo}[2]{#2}
\providecommand{\BIBentrySTDinterwordspacing}{\spaceskip=0pt\relax}
\providecommand{\BIBentryALTinterwordstretchfactor}{4}
\providecommand{\BIBentryALTinterwordspacing}{\spaceskip=\fontdimen2\font plus
\BIBentryALTinterwordstretchfactor\fontdimen3\font minus
  \fontdimen4\font\relax}
\providecommand{\BIBforeignlanguage}[2]{{%
\expandafter\ifx\csname l@#1\endcsname\relax
\typeout{** WARNING: IEEEtran.bst: No hyphenation pattern has been}%
\typeout{** loaded for the language `#1'. Using the pattern for}%
\typeout{** the default language instead.}%
\else
\language=\csname l@#1\endcsname
\fi
#2}}
\providecommand{\BIBdecl}{\relax}
\BIBdecl

\bibitem{goodfellow2016deep}
I.~Goodfellow, Y.~Bengio, and A.~Courville, \emph{Deep learning}.\hskip 1em
  plus 0.5em minus 0.4em\relax MIT Press, 2016.

\bibitem{zhang2016deep}
L.~Zhang, L.~Zhang, and B.~Du, ``Deep learning for remote sensing data: A
  technical tutorial on the state of the art,'' \emph{IEEE Geoscience and
  Remote Sensing Magazine}, vol.~4, no.~2, pp. 22--40, 2016.

\bibitem{DLforRS}
X.~X. {Zhu}, D.~{Tuia}, L.~{Mou}, G.-S. {Xia}, L.~{Zhang}, F.~{Xu}, and
  F.~{Fraundorfer}, ``{Deep learning in remote sensing: a review},''
  \emph{ArXiv e-prints}, Oct. 2017.

\bibitem{ronneberger2015u}
O.~Ronneberger, P.~Fischer, and T.~Brox, ``U-net: Convolutional networks for
  biomedical image segmentation,'' in \emph{International Conference on Medical
  Image Computing and Computer-Assisted Intervention}.\hskip 1em plus 0.5em
  minus 0.4em\relax Springer, 2015, pp. 234--241.

\bibitem{klambauer2017self}
G.~Klambauer, T.~Unterthiner, A.~Mayr, and S.~Hochreiter, ``Self-normalizing
  neural networks,'' \emph{arXiv preprint arXiv:1706.02515}, 2017.

\bibitem{clevert2015fast}
D.-A. Clevert, T.~Unterthiner, and S.~Hochreiter, ``Fast and accurate deep
  network learning by exponential linear units (elus),'' \emph{arXiv preprint
  arXiv:1511.07289}, 2015.

\bibitem{maggiori}
E.~Maggiori, Y.~Tarabalka, G.~Charpiat, and P.~Alliez, ``Convolutional neural
  networks for large-scale remote-sensing image classification,'' \emph{IEEE
  Transactions on Geoscience and Remote Sensing}, vol.~55, no.~2, pp. 645--657,
  Feb 2017.

\bibitem{audebert2016semantic}
N.~Audebert, B.~L. Saux, and S.~Lef{\`e}vre, ``Semantic segmentation of earth
  observation data using multimodal and multi-scale deep networks,''
  \emph{arXiv preprint arXiv:1609.06846}, 2016.

\bibitem{volpi2017dense}
M.~Volpi and D.~Tuia, ``Dense semantic labeling of subdecimeter resolution
  images with convolutional neural networks,'' \emph{IEEE Transactions on
  Geoscience and Remote Sensing}, vol.~55, no.~2, pp. 881--893, 2017.

\bibitem{holland1978role}
W.~R. Holland, ``The role of mesoscale eddies in the general circulation of the
  ocean—numerical experiments using a wind-driven quasi-geostrophic model,''
  \emph{Journal of Physical Oceanography}, vol.~8, no.~3, pp. 363--392, 1978.

\bibitem{chelton2011global}
D.~B. Chelton, M.~G. Schlax, and R.~M. Samelson, ``Global observations of
  nonlinear mesoscale eddies,'' \emph{Progress in Oceanography}, vol.~91,
  no.~2, pp. 167--216, 2011.

\bibitem{mcwilliams2008nature}
J.~C. McWilliams, ``The nature and consequences of oceanic eddies,''
  \emph{Ocean Modeling in an Eddying Regime}, pp. 5--15, 2008.

\bibitem{le2011parameterization}
J.~Le~Sommer, F.~d’Ovidio, and G.~Madec, ``Parameterization of subgrid
  stirring in eddy resolving ocean models. part 1: Theory and diagnostics,''
  \emph{Ocean Modelling}, vol.~39, no.~1, pp. 154--169, 2011.

\bibitem{faghmous2015daily}
J.~H. Faghmous, I.~Frenger, Y.~Yao, R.~Warmka, A.~Lindell, and V.~Kumar, ``A
  daily global mesoscale ocean eddy dataset from satellite altimetry,''
  \emph{Scientific data}, vol.~2, 2015.

\bibitem{pascual2006improved}
A.~Pascual, Y.~Faug{\`e}re, G.~Larnicol, and P.-Y. Le~Traon, ``Improved
  description of the ocean mesoscale variability by combining four satellite
  altimeters,'' \emph{Geophysical Research Letters}, vol.~33, no.~2, 2006.

\bibitem{faghmous2012eddyscan}
J.~H. Faghmous, L.~Styles, V.~Mithal, S.~Boriah, S.~Liess, V.~Kumar,
  F.~Vikeb{\o}, and M.~dos Santos~Mesquita, ``Eddyscan: A physically consistent
  ocean eddy monitoring application,'' in \emph{Intelligent Data Understanding
  (CIDU), 2012 Conference on}.\hskip 1em plus 0.5em minus 0.4em\relax IEEE,
  2012, pp. 96--103.

\bibitem{okubo1970horizontal}
A.~Okubo, ``Horizontal dispersion of floatable particles in the vicinity of
  velocity singularities such as convergences,'' in \emph{Deep sea research and
  oceanographic abstracts}, vol.~17, no.~3.\hskip 1em plus 0.5em minus
  0.4em\relax Elsevier, 1970, pp. 445--454.

\bibitem{weiss1991dynamics}
J.~Weiss, ``The dynamics of enstrophy transfer in two-dimensional
  hydrodynamics,'' \emph{Physica D: Nonlinear Phenomena}, vol.~48, no. 2-3, pp.
  273--294, 1991.

\bibitem{chelton2007global}
D.~B. Chelton, M.~G. Schlax, R.~M. Samelson, and R.~A. de~Szoeke, ``Global
  observations of large oceanic eddies,'' \emph{Geophysical Research Letters},
  vol.~34, no.~15, 2007.

\bibitem{turiel2007wavelet}
A.~Turiel, J.~Isern-Fontanet, and E.~Garc{\'\i}a-Ladona, ``Wavelet filtering to
  extract coherent vortices from altimetric data,'' \emph{Journal of
  Atmospheric and Oceanic Technology}, vol.~24, no.~12, pp. 2103--2119, 2007.

\bibitem{sadarjoen1999geometric}
I.~A. Sadarjoen and F.~H. Post, ``Geometric methods for vortex extraction,'' in
  \emph{Data Visualization’99}.\hskip 1em plus 0.5em minus 0.4em\relax
  Springer, 1999, pp. 53--62.

\bibitem{yi2014enhancing}
J.~Yi, Y.~Du, Z.~He, and C.~Zhou, ``Enhancing the accuracy of automatic eddy
  detection and the capability of recognizing the multi-core structures from
  maps of sea level anomaly,'' \emph{Ocean Science}, vol.~10, no.~1, pp.
  39--48, 2014.

\bibitem{isern2003identification}
J.~Isern-Fontanet, E.~Garc{\'\i}a-Ladona, and J.~Font, ``Identification of
  marine eddies from altimetric maps,'' \emph{Journal of Atmospheric and
  Oceanic Technology}, vol.~20, no.~5, pp. 772--778, 2003.

\bibitem{castellani2006identification}
M.~Castellani, ``Identification of eddies from sea surface temperature maps
  with neural networks,'' \emph{International journal of remote sensing},
  vol.~27, no.~8, pp. 1601--1618, 2006.

\bibitem{hai2008automatic}
J.~Hai, Y.~Xiaomei, G.~Jianming, and G.~Zhenyu, ``Automatic eddy extraction
  from sst imagery using artificial neural network,'' \emph{The international
  archives of the photogrammetry, remote sensing and spatial information
  science}, pp. 279--282, 2008.

\bibitem{ashkezari2016oceanic}
M.~D. Ashkezari, C.~N. Hill, C.~N. Follett, G.~Forget, and M.~J. Follows,
  ``Oceanic eddy detection and lifetime forecast using machine learning
  methods,'' \emph{Geophysical Research Letters}, vol.~43, no.~23, 2016.

\bibitem{deepeddy}
D.~Huang, Y.~Du, Q.~He, W.~Song, and A.~Liotta, ``Deepeddy: A simple deep
  architecture for mesoscale oceanic eddy detection in sar images,'' in
  \emph{2017 IEEE 14th International Conference on Networking, Sensing and
  Control (ICNSC)}, May 2017, pp. 673--678.

\bibitem{mason2014new}
E.~Mason, A.~Pascual, and J.~C. McWilliams, ``A new sea surface height--based
  code for oceanic mesoscale eddy tracking,'' \emph{Journal of Atmospheric and
  Oceanic Technology}, vol.~31, no.~5, pp. 1181--1188, 2014.

\bibitem{mason17}
E.~Mason, A.~Pascual, P.~Gaube, S.~Ruiz, J.~L. Pelegrí, and A.~Delepoulle,
  ``Subregional characterization of mesoscale eddies across the brazil-malvinas
  confluence,'' \emph{Journal of Geophysical Research: Oceans}, vol. 122,
  no.~4, pp. 3329--3357, 2017.

\bibitem{zeiler2010deconvolutional}
M.~D. Zeiler, D.~Krishnan, G.~W. Taylor, and R.~Fergus, ``Deconvolutional
  networks,'' in \emph{Computer Vision and Pattern Recognition (CVPR), 2010
  IEEE Conference on}.\hskip 1em plus 0.5em minus 0.4em\relax IEEE, 2010, pp.
  2528--2535.

\bibitem{milletari2016v}
F.~Milletari, N.~Navab, and S.-A. Ahmadi, ``V-net: Fully convolutional neural
  networks for volumetric medical image segmentation,'' in \emph{3D Vision
  (3DV), 2016 Fourth International Conference on}.\hskip 1em plus 0.5em minus
  0.4em\relax IEEE, 2016, pp. 565--571.

\bibitem{chollet2015keras}
F.~Chollet \emph{et~al.}, ``Keras,'' \url{https://github.com/fchollet/keras},
  2015.

\bibitem{kingma2014adam}
D.~Kingma and J.~Ba, ``Adam: A method for stochastic optimization,''
  \emph{arXiv preprint arXiv:1412.6980}, 2014.

\bibitem{he2015delving}
K.~He, X.~Zhang, S.~Ren, and J.~Sun, ``Delving deep into rectifiers: Surpassing
  human-level performance on imagenet classification,'' in \emph{Proceedings of
  the IEEE international conference on computer vision}, 2015, pp. 1026--1034.

\end{thebibliography}
\end{document}